# Exploring the Plausibility of Hate and Counter Speech Detectors with Explainable AI

Adrian Jaques Böck
Adrian.Boeck@fhstp.ac.at
Institute of Creative Media
Technologies
St. Pölten, Austria

Djordje Slijepčević
Djordje.Slijepcevic@fhstp.ac.at
Institute of Creative Media
Technologies
St. Pölten, Austria

Matthias Zeppelzauer
Matthias.Zeppelzauer@fhstp.ac.at
Institute of Creative Media
Technologies
St. Pölten, Austria

## ABSTRACT

In this paper we investigate the explainability of transformer models and their plausibility for hate speech and counter speech detection. We compare representatives of four different explainability approaches, i.e., gradient-based, perturbation-based, attention-based, and prototype-based approaches, and analyze them quantitatively with an ablation study and qualitatively in a user study. Results show that perturbation-based explainability performs best, followed by gradient-based and attention-based explainability. Prototype-based experiments did not yield useful results. Overall, we observe that explainability strongly supports the users in better understanding the model predictions.

## 1 INTRODUCTION

A prevalent problem today on social media platforms like Instagram, TikTok and X is hate speech (HS) and toxic content [17]. Mechanisms like reporting and banning have shown to be ineffective in many cases and manual content moderation is time-consuming, expensive and straining for the moderators [10]. One promising solution to fight online hate is *counter speech* (CS), i.e., actively countering hateful messages [4]. CS is fast, direct and context-sensitive, preserves the freedom of speech, and has the potential to mobilize other users to show online courage, thereby facilitating sustainable changes in communication culture [15]. CS appears in different forms, e.g., as presented facts, factual counter-argumentation, humorous messages, cynical messages, as well as in the form of counter hate (shouting back) [5].

The automated detection of CS is a complex task due to its sophisticated semantics, the different types, and the often missing contextual information required to classify content as CS [44]. Recently, first methods have emerged for the automated detection and classification of CS, accompanied with labeled CS datasets [18, 25, 29, 35, 44, 45]. First results show the potential of natural language processing (NLP) models for this task. A key limitation of today's NLP models, however, is that they are opaque in their decision-making process and in their internal workings. The complex computational mechanisms which drive the outputs of today's NLP models (esp. transformer architectures) are difficult to interpret by humans. Thus, given a certain prediction, it is not clear how this prediction is grounded in the input data and to what degree the model has understood the semantics of the content. This limits the usability of the approaches and the trust placed in them.

In this paper, we want to shed light on the internal workings and the learned problem understanding of automated CS detectors by leveraging different explainability approaches [34]. To this end, we investigate three main research questions:

- RQ1: How well can we solve the problem of HS/CS classification with recent architectures for language modeling?
- RQ2: Do the trained models ground their decisions on meaningful information?
- RQ3: Are explainability mechanisms for NLP models useful to explain the model behavior and grounding and which explainability approach is most helpful for humans in interpreting the results?

To answer our research questions, we train transformer models to solve classification tasks on public HS and CS related datasets. Subsequently, we explore four distinct approaches to explainability, including gradient-based, perturbation-based, attention-based, and prototype-based methods. We conduct experiments including a representative method from each of the four explainability approaches: i.e., Local Interpretable Model-agnostic Explanations (LIME) [36], Integrated Gradients (IG) [40], GlobEnc [32], and ProtoTEx [11]. We perform classification experiments, a quantitative ablation study as well as a qualitative user study to investigate our research questions.

The main contributions of our work are: (i) a systematic comparison of several principally different XAI approaches for explaining

NLP models including quantitative measures and human judgements on faithfulness; (ii) an analysis of the plausibility, understandability, sufficiency, trustworthiness, satisfaction, and helpfulness of different XAI methods for humans; and (iii) a first investigation on the explainability and grounding of NLP models for detecting counter speech.

## 2 RELATED WORK

Deep neural networks, such as transformer models, exhibit challenges regarding interpretability and transparency. The importance of understanding the decision-making process of such models, especially when analyzing sensitive content, has stimulated recent efforts in the field of Explainable AI (XAI) [3, 6, 7, 23, 43]. Despite numerous studies comparing XAI methods for text-based models [3, 23, 24, 43], there is a lack of a systematic comparison for HS and CS detection. This section provides an overview of XAI approaches utilized in the NLP field for text classification with an emphasis on transformer models. Table 1 outlines XAI methods for NLP models and corresponding text classification tasks. For a general review of visualization and XAI methods for the transformer model BERT [12], refer to the survey conducted in [7].

Attanasio et al. [3] investigated the explainability of BERT in the context of detecting misogyny in tweets. The authors compared four feature attribution methods, including Gradients, IG, SHapley Additive exPlanations (SHAP), and Sampling and Occlusion (SOC). Plausibility and faithfulness metrics revealed an inconsistency and lack of faithfulness in gradient-based methods. SHAP and SOC were found to be more faithful and plausible.

Wu and Ong [43] focused on sentiment classification with BERT, evaluating attribution methods like Gradient Sensitivity (GS), Gradient*Input (GI), Layer-wise Relevance Propagation (LRP), and Layer-wise Attention (LAT). Ablation studies assessed validity, revealing consistent changes in GS, GI, and LRP, while LAT was less reliable.

Bodria et al. [6] trained BERT for sentiment analysis and evaluated IG, LIME, and attention scores obtained via an additional attention layer. They evaluated the fidelity of the XAI methods by comparing BERT's predictions with the explanation scores, i.e., by determining the algebraic sign of the sum of explanation scores within a sentence. The results showed that IG exhibited the highest fidelity, followed by LIME, and attention scores.

Krishna et al. [23] trained an LSTM to compare six post-hoc XAI methods. The results revealed disagreements in heuristic evaluations and user studies. They suggested the need for standardized evaluation metrics and the inclusion of user preferences in the selection of XAI methods.

Velampalli et al. [41] trained different sentiment analysis models and utilized SHAP for explanation. SHAP effectively identified model strengths and weaknesses but lacked in quantitative evaluation metrics.

Ansari et al. [2] enhanced their hate speech detection based on LSTM and CNN through data augmentation and explained the models using LIME and IG. The proposed augmentation approach led to an improvement of the models and explanations. However, no systematic evaluation of the XAI methods was performed.

Sebbaq and Faddouli [37] proposed MTBERT-Attention, an attention-based XAI method. The attention scores are obtained by training an additional co-attention layer alongside the BERT encoder that computes attention tokens for each word in the dataset. The attention scores are aggregated per word and are then utilized in a simple (explainable) softmax classifier to predict the labels. While achieving high fidelity, their approach overrated frequently adverbs and prepositions. LIME, which served as a baseline, focuses on local explanations that provide more detailed insights for specific instances.

Mehta and Passi [31] explored different BERT variants and LIME for hate speech detection. They used ERASER [13], which is a benchmark for measuring the explainability of NLP models by evaluating how well models align with human rationales, assessing plausibility and faithfulness.

Das et al. [11] introduced ProtoTEx, a white-box model for propaganda detection based on prototype networks and a underlying BART encoder [26]. The approach involves generating prototype tensors from encoded latent clusters derived from training samples, utilizing transformer encoders. ProtoTEx matches the performance of the BART-large [26] model and outperforms BERT-large [12] and KNN-BART baselines. The authors noted that the prototypes might not be always beneficial for explaining the predictions, especially if the prototypes are closely associated with only a few training samples.

Sourati et al. [39] extended the ProtoTEx method from binary to multi-class classification and replaced the original BART encoder with Electra [9]. However, they encountered difficulties with inconsistent prototypes, potentially stemming from the challenging nature of the fallacy identification task.

An overall conclusion from the related work suggests that identifying the most suitable method for explaining transformer models relies on the particular use case and underlying dataset. There is no clear winner among explainability approaches, neither for global nor for local explanations. LIME and IG are state-of-the-art XAI methods frequently used as reference methods in the related work, thus they are incorporated in the evaluation in this paper. Additionally, we incorporate two more recent and complementary XAI methods in our experiments, i.e., GlobEnc and ProtoTEx, as representatives for attention-based and prototype-based approaches. Unlike the mentioned studies that focus on limited sets of performance metrics, we evaluate a broader range of metrics, offering a more comprehensive and comparative analysis.

## 3 METHODS

### 3.1 Datasets

We employ two publicly available CS-related datasets. The **Thou Shalt Not Hate (TSNH) dataset** [30] contains English comments from YouTube, with an almost balanced distribution of 7,024 CS and 6,898 non-CS samples, totaling 13,922 samples. The training set comprises 8,909 samples, the validation set 2,228, and the test set 2,785 samples. The **HateCounter (HC) dataset** [29] consists of pairs of a tweet and a reply sourced from X (Twitter), encompassing 1,290 pairs of HS and CS as well as 223 pairs of HS and HS supporting replies. Only the replies were used in this work, which results in a highly imbalanced dataset. The dataset is divided into a training set of 1,059 samples, a validation set of 302, and a test set of 152 samples. Both datasets were pre-processed by replacing links and user references with "<LINK>" and "<USER>".

Table 1: Summary of explainability methods used in the related work.

| Author | Use Case | Model / Explainability Method |
| --- | --- | --- |
| Attanasio et al. [3] | Misogyny Detection | BERT / Gradients, IG, SHAP, SOC |
| Wu and Ong [43] | Sentiment Analysis | BERT / GS, GI, LRP, LAT |
| Bodria et al. [6] | Sentiment Analysis | BERT / IG, LIME, Attention Scores |
| Krishna et al. [23] | News Article Classification | LSTM / LIME, KernelSHAP, SmoothGrad, GI, IG, GradCAM |
| Velampalli et al. [41] | Sentiment Analysis | SBERT and USE + FCNN and LSTM / SHAP |
| Ansari et al. [2] | Hate Speech Detection | LSTM, CNN / LIME, IG |
| Sebbaq and Faddouli [37] | Cognitive Text Classification | MTBERT-Attention (self-explaining classifier), LIME |
| Mehta and Passi [31] | Hate Speech Detection | Decision Tree, LSTM, BERT (different variants) / LIME |
| Das et al. [11] | Propaganda Detection | BERT, BART, KNN-BART / ProtoTEx |
| Sourati et al. [39] | Local Fallacy Identification | Electra, BERT, DeBERTa, RoBERTa, DistilBERT / ProtoTEx, Case-based Reasoning Framework, Knowledge Injection Framework |

## 3.2 Classification Models

For the experiments, we utilized a pre-trained BERT model (bert-base-uncased) from Hugging Face [21]. We fine-tuned the BERT model separately on the training data of the two datasets, i.e., resulting in one BERT model for the TSNH dataset and a second BERT model for the HC dataset. For fine-tuning, we employed a training batch size of 8, an evaluation batch size of 16, a learning rate of 3e-5, and the AdamW optimizer [28]. The training process was restricted to a maximum of 15 epochs to mitigate the risk of overfitting. Additionally, an early stopping mechanism with a patience of 5 was applied to the validation loss.

## 3.3 Explainability Methods

**IG** [40] is an extension of traditional gradient-based approaches. Traditional approaches compute the partial derivative of the output with respect to each input feature, thereby assessing how slight variations in the input influence the prediction. This process is, however, sensitive to noise and non-linearities and not always reliable [1]. IG is more robust than other gradient-based methods in addressing the non-linear behavior of neural networks by computing the gradient not just for the input vector but for various variants, starting with a baseline (usually neutral, such as zero values) and progressively moving towards the actual input vector. During this process, IG calculates the average gradient and the contribution of each feature to the model's prediction. IG is applicable to various network architectures and easy to use. However, non-linearity and high-dimensional input vectors may introduce noise, leading to attribution maps focusing on irrelevant features [1].

**LIME** [36], is a post-hoc and model-agnostic XAI method for explaining complex model predictions. It utilizes locally linear surrogate models, created by training on various perturbations generated by removing or altering parts of the input data. This method produces local explanations by approximating the predictions of the original model with the use of a simpler surrogate model. Despite its computational intensity and sensitivity to input changes, LIME's capability to work with any classification or regression model without accessing internal weights or parameters makes it a popular XAI method.

**GlobEnc** [32] is a global token attribution method for transformer models. It does not just look at the attention mechanisms but considers almost every part of the encoder layers of the model. The computation of the attribution scores (which can be seen as feature attributions) is done by a modified attention roll-out technique. It considers not only the attention weights but also the magnitude of the vectors involved, the residual connections, and layer normalization effects, across all layers of the transformer. This provides comprehensive attributions, showing high correlation with gradient-based methods, when considering the entire model's behavior and interactions across all its components. According to [32], it is robust and computationally efficient.

**ProtoTEx** [11] is based on case-based reasoning that combines pre-trained language model encoders with a linear prototype layer for creating prototype tensors. These tensors are formed by the model identifying and grouping together similar training examples into latent clusters. This process involves calculating the Euclidean distance between each input and the prototypes within the layer, thereby creating a representation space shared by the input data and the prototype layer. Through this, the ProtoTEx model effectively categorizes input sentences by comparing them to these prototypes, facilitating a transparent and interpretable (i.e., white-box) classification process.

## 4 EVALUATION

### 4.1 Classification Performance

The performance of the trained HS and CS detectors is evaluated on the test set of the respective dataset to obtain a performance baseline. The evaluation encompassed several key metrics including the accuracy, recall (macro), precision (macro), and $F_1$ score (macro). The performance evaluation is crucial to ensure that the models are sufficiently robust for analyzing the XAI methods.

### 4.2 Ablation Study

The goal of the ablation study is to investigate the faithfulness [22] of the methods. Our assumption is that removing the token(s) that the XAI method considers most important for a certain class will reduce the likelihood of the class. The more tokens we remove, the stronger this effect should be. The ablation study thereby serves as a sanity check to evaluate whether the XAI methods work as expected. Note that, removing tokens can also introduce a bias as perturbed inputs can differ in their nature compared to complete inputs as indicated in [16].

In the ablation study, we calculate and sort the attribution scores of each XAI method in descending order for all input tokens. We focus on the top four tokens with the highest scores, since we consider them as most influential. Iteratively, each of these tokens is removed, and the resulting class likelihood is compared to that without token removal. A decrease in likelihood indicates a performance drop.

For these experiments we only considered sentences with 10 to 30 tokens to ensure that the sentences exhibit a certain degree of expressiveness. As a result, for the TSNH dataset 1040 test sentences were selected. For the HC dataset, all 57 test sentences were selected. Next, we selected only sentences with at least four positive attribution scores across all investigated methods. These adjustments resulted in a final count of 535 sentences for TSNH and 55 for the HC dataset for the ablation study. Note that for GlobEnc we took the attribution scores from the last layer of the BERT transformer.

### 4.3 User Study

We conducted a user study to qualitatively assess criteria including the confidence of the participants using the XAI methods, plausibility [22], understandability and trustworthiness [27, 33], sufficiency [8, 19], satisfaction [29], and helpfulness [29]. The user study encompassed two tasks: a forward prediction task and a comparative study. The user study was conducted using an online questionnaire via LimeSurvey where users were asked to assess the above-listed criteria. Only text from the TSNH dataset was used since it consists of significantly more samples than the HC dataset. The resources of the user study are published on GitHub at https://github.com/JaquJaqu/CBMI_XAI.

**Task 1: Forward Prediction.** Participants were asked to guess the model's prediction based on the textual content of the input sentence and the explanations provided by the different XAI methods to assess the plausibility of the methods. Participants were unaware of the utilized XAI method and the model's actual prediction. A set of five sentences from the TSNH dataset was selected, which were correctly classified by the underlying classification model. This set was kept small to reduce the time effort of the study. The same sentences were used for all questions of this task of the survey to assure comparability. The study started with a baseline assessment of the textual content without additional explanations to assess the participants' intuitive predictive abilities. Subsequently, the participants evaluated the content with additional explanations (XAI methods selected in blind fashion and in random order). Special tokens added by the tokenizers were removed. Participants provided the confidence in their predictions using a Likert scale ranging from 1 to 5.

Attribution scores were normalized and visualized via text background color using the colormap illustrated in Figure 1. The colors green and violet were selected to represent positive and negative attributions of the tokens, respectively. This design ensures that the participants are not able to infer the specific XAI method based on the visualization. Their shades vary depending on the magnitude of the attribution scores, with greater positive or negative values being represented by more intense coloration.

no you are wrong there are over 70 sects in islam which one are you talking about

**Figure 1: CS example of an LIME explanation with color-coded attribution scores (green for positive and violet for negative).**

**Task 2: Comparative Study.** Participants were asked to rate five criteria on a Likert scale from 1 to 5: understandability, sufficiency, trustworthiness, satisfaction, and helpfulness. Participants were given five texts, different from those in the first task, and were provided with both the ground truth and the predicted labels for each text. Unlike in the previous task, participants were informed that they were analyzing explanations from different XAI methods. The participants were shown the original explanation of the XAI methods (the original visualization as proposed by the authors of the methods) to solve the task. For each evaluation criteria, we investigated differences in the distributions between the XAI methods. To compare the XAI methods, we chose to employ the non-parametric Friedman test [14], as the data did not follow a normal distribution. In the case the output of this test was statistically significant ($p < 0.05$), we performed pairwise comparisons of the post-hoc analysis using the non-parametric Wilcoxon signed-rank test [42]. We applied the Holm-Bonferroni [20] correction to account for the family-wise error rate.

## 5 RESULTS AND DISCUSSION

### 5.1 Classification Results

We first evaluate the two fine-tuned BERT models trained on the TSNH and HC datasets to obtain a performance baseline. To provide a basis for comparison we compute the random baseline for each dataset, which is 0.50 for TSNH and 0.45 for the HC dataset. Table 2 shows the performance of the fine-tuned BERT classifiers.

**Table 2: Classification results for the test data including accuracy, precision (macro), recall (macro), and $F_1$ score (macro).**

| Model | Data | Acc. | Precision | Recall | $F_1$ Score |
|---|---|---|---|---|---|
| BERT | TSNH | 0.7268 | 0.7554 | 0.7230 | 0.7166 |
| BERT | HC | 0.8618 | 0.7153 | 0.6360 | 0.6610 |

Both BERT models exceeded the respective random baseline. BERT trained on the TSNH dataset achieves an accuracy of 72.7% and an $F_1$ score of 0.72. BERT trained on the HC dataset, achieves a high accuracy of 86.2% and an $F_1$ score of 0.66. Overall, these results indicate that both models are able to learn task-specific patterns, which is important when investigating XAI methods in their context. An objective assessment of explainability through XAI methods is only meaningful when the underlying classification model captures meaningful class-specific information. Unreliable models may introduce biases from unstable patterns, noise, and spurious correlations, which may in turn bias the explainability results [38].

## 5.2 Note on ProtoTEx Results

The evaluation of the ProtoTEx method involved training a model utilizing a BERT encoder on the TSNH dataset. The results reveal a high sensitivity to the number of prototypes, which had a significant impact on model performance. The achieved accuracy on the test set was 0.52%, which is close to the random baseline of 0.50%. In addition to the very poor performance, which prevents a meaningful evaluation of explainability, the method's explanations were counter-intuitive (e.g., providing prototypes as explanations originating from a class other than the predicted one). This behavior was also observed in previous studies [11, 39]. Due to the low classification performance and noisy explanations, we excluded ProtoTEx from further analysis and comparison.

## 5.3 Results of the Ablation Study

Figure 2 (top) shows the changes in prediction probabilities over all ablations for the TSNH and HC datasets, respectively. Our expectation is that removing tokens with high (positive) attribution scores should generally result in a negative change (drop) in class likelihood. For both datasets and all three XAI methods the median change in class likelihood is around zero, with the HC dataset showing a stronger tendency towards negative changes.

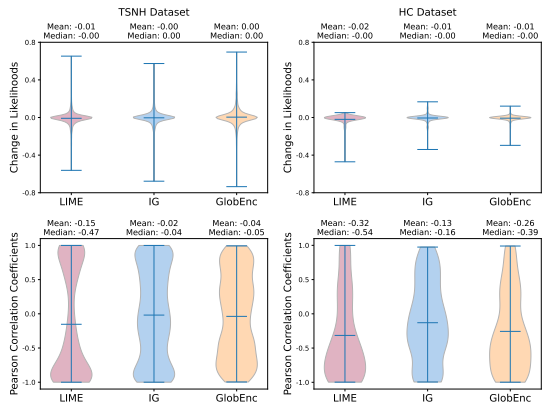

**Figure 2: Change in class likelihood and Pearson correlations (calculated on both original and ablated data) with the attribution scores of the ablated tokens across all three XAI methods for both datasets, i.e., TSNH (left) and HC (right).**

Figure 4 illustrates a more comprehensive analysis of the changes in class likelihood among the individual top-ranked tokens. There is a clear pattern: the largest change in class likelihood occurs when removing the most important token according to the XAI methods (i.e., the one with the highest attribution score). This change gradually decreases for the subsequent tokens. Only minor changes in class likelihood are observed for token 4. Notably, only LIME shows a more pronounced decrease in predicted class likelihoods, particularly evident for the HC dataset and thereby best fulfills the expected behavior. For the other two XAI methods, we observe both decreases and increases of class likelihoods when removing tokens with high attribution scores. This is an unexpected behavior which questions the faithfulness of the methods.

The visualizations of the Pearson correlation between the attribution score of a token and the change in class likelihood observed when removing the token in Figure 2 (bottom) show different trends across the XAI methods and both datasets. In the case of the TSNH dataset with LIME, we can observe a bimodal distribution, indicating that high attribution scores are accountable for high changes in class likelihoods, with a greater number of samples exhibiting higher attribution being strongly correlated with larger decreases in the class likelihood. In contrast, the correlations for IG and GlobEnc appear more uniform, indicating that the entire correlation spectrum is represented across all test samples. In the case of the HC dataset, all XAI methods demonstrate predominantly negative correlations, i.e., the most relevant tokens lead to a drop in class likelihood (as expected for a well-working XAI method). Notably, LIME exhibits the strongest overall correlation, with a median of -0.54, followed by GlobEnc with a median of -0.39 for the HC dataset. IG shows slightly different results with a median of -0.16, indicating a negligible correlation compared to the other two methods.

Overall, for both datasets, the findings from the ablation study indicate that LIME produces the most plausible results compared to IG and GlobEnc. These findings are partly supported by Bodria et al. [6], who also confirmed higher fidelity with LIME compared to attention-based XAI methods. However, they observed that IG exhibited the highest fidelity in their study.

Generally, the decrease in class likelihood resulting from the ablations in our study is lower than initially expected. This could be due to various reasons. One possibility is that we are removing tokens rather than entire words, thus certain parts of the words that remain may still contribute to the class likelihood and thus dampen the effect. Furthermore, even if the tokens represent entire words, in certain scenarios their semantic meaning (that is important for the task) is maintained by other words in the sentence (e.g., with multiple swearwords, the removal of one may result in minimal changes in the predicted likelihood).

## 5.4 User Study Results

**Task 1: Forward prediction.** Figure 5 (left) presents the user study results in which participants were asked to guess the model's prediction based on the textual content only, as well as with explanations provided by the XAI methods LIME, IG, and GlobEnc. There is a difference of at least 12.9% in the classification accuracy of the predictions made by the study participants when they receive explanations in addition to the textual content. The results with the additional explanations are similar to each other, with explanations by IG leading to 2.9% better performance compared to the other two methods. Interestingly, these trends in the classification results do not reflect in the self-assessment of confidence (Figure 5 (right)). The confidence in the self-made predictions was highest within the baseline approach without explanations (mean of 3.6), closely followed by GlobEnc (mean of 3.5). Third place is LIME (mean of 3.0) followed by IG (mean of 2.7). Interestingly, IG achieve the lowest confidence but effectively provided the most benefit to the users in the prediction task.

**Task 2: Comparative Study.** The results for the five different evaluation criteria are depicted in Figure 3. Each subfigure visualizes the results for all three XAI methods for a certain evaluation criterion. Across all criteria, LIME achieved the top average scores (ranging from 2.7 to 3.4), followed by IG (ranging from 2.5 to 3.0), and last by GlobEnc (ranging from 2.0 to 2.8).

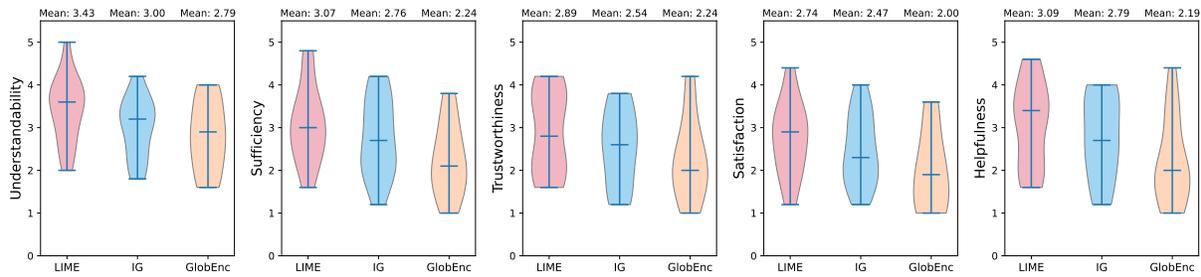

Figure 3: Comparison of Likert scores of the three XAI methods across the five criteria: understandability, sufficiency, trustworthiness, satisfaction, and helpfulness.

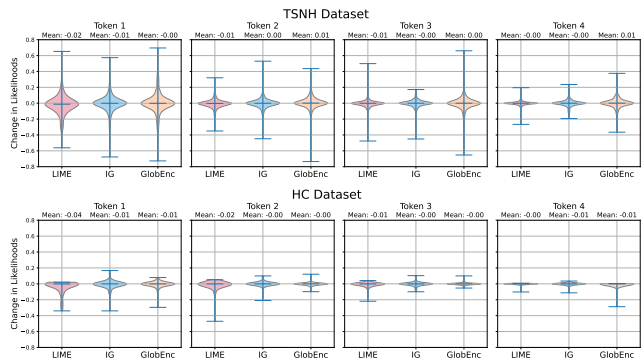

Figure 4: Change in class likelihood for all three XAI methods and all four ablated tokens for both datasets.

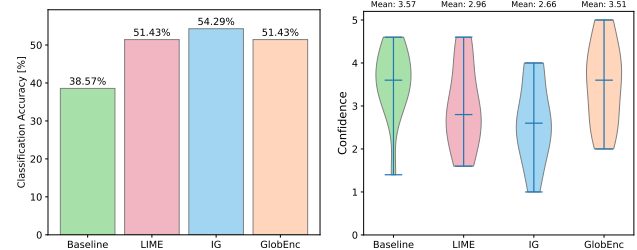

Figure 5: Prediction accuracies of participants using only textual content (baseline) and with additional explanations (left). Comparison of Likert scores reflecting participants' confidence levels (right).

The results of the Friedman tests show a significant difference among the three XAI methods across all criteria, with following $p$-values: $p = 0.0034$ for understandability, $p = 0.0002$ for sufficiency, $p = 0.0013$ for trustworthiness, $p = 0.0006$ for satisfaction, and $p < 0.0001$ for helpfulness. A post-hoc analysis (see Table 3) using pairwise Wilcoxon signed-rank tests with Holm-Bonferroni adjustment shows that LIME explanations were perceived as superior to GlobEnc explanations across all criteria, with statistical significance ($p < 0.05$). Except for satisfaction, LIME explanations were also perceived as superior to IG explanations across all criteria, with statistical significance ($p < 0.05$). Across all criteria, except for understandability, IG explanations were perceived as superior to GlobEnc explanations, with statistical significance ($p < 0.05$).

|  | LIME vs. IG | LIME vs. GlobEnc | IG vs. GlobEnc |
|---|---|---|---|
| Understandability | 0.0492* | 0.0007* | 0.1708 |
| Sufficiency | 0.0281* | 0.0001* | 0.0015* |
| Trustworthiness | 0.0283* | 0.0004* | 0.0475* |
| Satisfaction | 0.0519 | 0.0004* | 0.0132* |
| Helpfulness | 0.0493* | 0.0001* | 0.0031* |

Table 3: Results of post-hoc analysis on the user study data using Wilcoxon signed-rank tests with Holm-Bonferroni adjustment. An asterisk symbol (*) indicates a statistically significant difference between the XAI methods ($p < 0.05$).

## 6 CONCLUSION

This paper examined strengths and limitations of different XAI approaches, i.e., LIME, IG, GlobEnc, and ProtoTEx for the explanation of transformer-based classification models for hate and counter speech. The classification models demonstrate decent performance in classification (except for the ProtoTEx model) ranging from 73% to 86% in accuracy (RQ1). The results of the ablation study demonstrate that LIME consistently showed more impact on class likelihoods upon token removal and thus more faithful results in comparison to IG and GlobEnc (RQ3). The results of the user study further showed that LIME outperforms IG and GlobEnc across most metrics, i.e., understandability, sufficiency, trustworthiness, and helpfulness (RQ3). Another finding from the user study is that XAI methods strongly assist the users in predicting the model's output, which indicates that their explanations are meaningful for the task, speaking also for a reasonable grounding of the models (RQ2). Interestingly, the XAI methods did not contribute to the confidence of the users in their predictions. This is unexpected and will be subject to further research.

## ACKNOWLEDGMENTS

This work has been funded by the Vienna Science and Technology Fund (WWTF) [10.47379/ICT20016] and the Austrian Research Promotion Agency (FFG) under Grant no. 898085.